# Modeling Uncertain Temporal Evolutions in Model-Based Diagnosis

**Luigi Portinale**
Dipartimento di Informatica - Universita' di Torino
C.so Svizzera 185 - 10149 Torino (Italy)

## Abstract

Although the notion of diagnostic problem has been extensively investigated in the context of static systems, in most practical applications the behavior of the modeled system is significantly variable during time. The goal of the paper is to propose a novel approach to the modeling of uncertainty about temporal evolutions of time-varying systems and a characterization of model-based temporal diagnosis. Since in most real world cases knowledge about the temporal evolution of the system to be diagnosed is uncertain, we consider the case when probabilistic temporal knowledge is available for each component of the system and we choose to model it by means of Markov chains. In fact, we aim at exploiting the statistical assumptions underlying reliability theory in the context of the diagnosis of time-varying systems. We finally show how to exploit Markov chain theory in order to discard, in the diagnostic process, very unlikely diagnoses.

## 1 INTRODUCTION

The notion of diagnostic reasoning, and in particular of model-based diagnosis, has been extensively investigated in the past and two basic paradigms emerged: the *logic* and the *probabilistic* one. In the logic paradigm it is often the case that the model represents the function and structure of a device; in this case the model usually represents the normal behavior of a system and the diagnostic reasoning follows a *consistency-based approach* [21] (the diagnosis is a set of abnormality assumptions on the components of the system that, together with the observations, restore the consistency in the model). Alternatively, the use of *causal models* has been proposed to model the faulty behavior of the system where an *abductive approach* is used [19] (the diagnosis must cover the observations). Both approaches use a logical model of the system to be diagnosed and it has been shown that they are just the extremes of a spectrum of logical definitions of diagnosis [6]. On the other hand, in the probabilistic paradigm, diagnostic knowledge is usually represented by means of associations between symptoms and disorders; performing a diagnosis means finding the most probable set of disorders, given the symptoms. The major part of the work in probabilistic diagnosis has been concerned with the use of Bayesian methods based on *belief networks* [17] as in MUNIN [1], in QMR-BN [22] and in [12]. However, other approaches have been proposed, for instance, relying on Dempster-Shafer theory [13] or on the integration of Bayesian classification with the set covering model [18].

These two basic paradigms to diagnosis seem to be complementary for certain aspects (for example, probabilistic diagnostic reasoning well addresses the problem of minimizing the cost of the computation, while logic-based reasoning is in general more flexible in dealing with multiple disorders or contextual information); for this reason some attempts have been made for combining them. In fact, probabilistic information has been integrated in logical diagnostic framework either for defining criteria for choosing the best next measurement [5,8] or for extending probabilistic diagnostic frameworks to non propositional form [20]. However, the context in which such approaches have been proposed is mainly that of static systems in which the system behavior can be thought as fixed during the diagnostic process and so time independent. Clearly, in important applications this is not sufficient, because the modeled system is intrinsi-cally dynamic and its behavior is significantly variable during time. For this reason there is a growing interest in the study of the behavior of time-varying systems in the attempt of either extending static diagnostic techniques [14] or finding new approaches more suitable for time-dependent behavior [11]. Diagnosing systems with time-varying behavior requires the ability of dealing at least with the following important aspects: observations across different time instants and explicit description of state changes of the system with respect to time. One of the immediate consequences of these aspects of the problem is the need of having some form of abstraction in the model representing the system to be diagnosed; in fact, although some of the earlier approaches to diagnosis of dynamic systems tried to deal with single level description of the system [24], most of the recent proposals are focused on the possibility of having a hierarchical model in which multiple layers of abstractions can be used in order to perform the diagnostic task [14,16].

In previous works we concentrated on abductive reasoning as an useful framework for diagnosing static systems, even if some attention has to be paid for mitigating the computational complexity of the abductive process [2]. However, extending these mechanisms in the direction of time-varying systems can lead to several problems. First of all, in most real world cases temporal information is uncertain and not always easy to encode in the model of the system to be diagnosed. The use of propagation techniques for fuzzy temporal intervals has been proposed in[4] and[10], but even if in practice we can determine the temporal evolutions of the system in a more precise way, serious problems arise from the computational point of view [7]; this means that some kind of approximation has to be considered.

In particular, we are interested in analyzing how to exploit, in a component-oriented model, probabilistic knowledge about possible temporal evolutions of the components. In particular we assume that:

- components have a discrete set of different *behavioral modes* [9] (a behavioral mode represents a particular state of a component; the set of behavioral modes consists of one "correct" mode and, in general, several abnormal modes representing faulty conditions of the component).
- each component can change its mode during time, while the manifestations of a behavioral mode are instantaneous (with respect to the amount of time required for a transition between two behavioral modes, or the amount of time elapsing between two consecutive time points with observations);
- knowledge about the temporal evolution of a component is uncertain and has to be modeled at some level of abstraction.

The aim of the paper is to propose, given this kind of assumptions, a characterization of temporal diagnosis in such a way that diagnostic techniques, developed for static systems, could be used as much as possible in the diagnosis of systems exhibiting time-varying behavior. This is pursued having in mind the fact that temporal information is abstracted from other behavioral features of the system, such as the relationships between behavioral modes of the components and their observable manifestations; this means that such relationships do not take into account time which is added to the model as an orthogonal dimension (see also [15]). The kind of abstraction we use throughout the paper is a probabilistic one. We assume to have at disposal *probabilistic knowledge* about the temporal behavior of the components of the system to be diagnosed in such a way that such behavior can be modeled as a stochastic process; in particular, we aim at exploiting the theory of markovian stochastic processes in a diagnostic setting by adopting the usual assumptions followed in *reliability theory* [23]. In fact, this kind of probabilistic knowledge can be supposed to be available from the statistics about the behavior of the system. We will show how this approach can be used in order to discard temporal evolutions which are very unlikely, however, for the lack of space, we discussed here just the declarative characterization of the problem, without considering reasoning issues.

## 2 STOCHASTIC PROCESSES AND RELIABILITY THEORY

In this section, we briefly recall some basic notions relative to stochastic processes and the probabilistic assumptions usually adopted in reliability theory, concerning the life cycle of a component of a physical system.

**Definition 1** *A* **stochastic process** *is a family of random variables* $\{X(t)/t \in T\}$ *defined over the same probability space, taking values in a set* $S$ *and indexed by a parameter* $t \in T$.

The values assumed by the variables of the stochastic process are called *states* and the set $S$ is called the *state space* of the process. Usually the parame-ter $t$ represents time, so a stochastic process can be thought as the model of the evolution of a system across time[1]. In the following we will concentrate on *chains* (i.e. *discrete-state processes*) with discrete time parameter and in particular on *Markov chains*.

**Definition 2** *A* **Markov chain** *is a discrete-state stochastic process* $\{X(t)/t \in T\}$ *such that for any* $t_0 < t_1 < ...t_n$ *the conditional distribution of* $X(t_n)$ *for given values* $X(t_0)\ldots X(t_{n-1})$ *depends only on* $X(t_{n-1})$ *that is:*

$$P[X(t_n) = x_n|X(t_{n-1}) = x_{n-1}\ldots X(t_0) = x_0] = P[X(t_n) = x_n|X(t_{n-1}) = x_{n-1}]$$

We will be concerned only with Discrete-Time Markov Chains (DTMC). In a Markov chain, the probability of transition from one state to another depends only from the current state and from the current time instant. If the conditional probability showed in the above definition is invariant with respect to the time origin, then we have a so called *time − homogeneous* Markov chain; this means that

$$P[X(t_n) = x_n|X(t_{n-1}) = x_{n-1}] = P[X(t_n - t_{n-1}) = x_n|X(0) = x_{n-1}]$$

In this case, the past history of the chain is completely summarized in the current state; with the assumption of time-homogeneity, it can be shown that, in the case of a DTMC, the sojourn time in a given state follows a *geometric distribution*. This is the only distribution satisfying the *memoryless property*, for a discrete random variable (the corresponding memoryless distribution for a continuous variable is the *exponential distribution*). A discrete random variable $X$ is geometrically distributed with parameter $p$ if its *probability mass function* is $pmf(t) = P(X = t) = p(1-p)^{t-1}$. We say that $X$ has the memoryless property if and only if $P(X = t + n|X > t) = P(X = n)$; this means that we need not remember how long the process has been spending time in a given state to determine the probabilities of the next possible transitions (i.e. we can arbitrarily choose the origin of the time axis). In the following we will consider only time-homogeneous DTMC.

Reliability theory copes with the application of particular probability distributions to the analysis of the life cycle of the components of a physical system. In particular, it has been recognized that the life of a component can be subdivided into three phases: the *infant mortality phase* where the *failure probability* $p$ is decreasing with time, the *usual life phase* with constant failure probability and the *wear-out phase* where the failure probability is increasing with age. If we consider the lifetime $X$ of a component to be a discrete random variable[2], in the usual life phase $X$ follows a geometric distribution with parameter $p$.

[1] In the following we will refer to $t$ as the time parameter.

## 3 DIAGNOSTIC FRAMEWORK

In the introduction, we discussed problems concerning the management of temporal information in diagnostic systems; one of the basic difficulty is relative to the approach to be followed in modeling such a kind of information. Since a lot of work in reliability theory has lead to well defined statistical assumptions about the temporal behavior of the components of a system, we aims at extending such assumptions in a model based diagnostic framework.

### 3.1 REPRESENTATIONAL ISSUES

Following the approach proposed in [3], we choose to decompose the model of the system to be diagnosed into two parts:

- **a logical static behavioral model** (with no representation of time);
- **a mode transition model** showing the possible temporal evolutions of the behavioral modes of each component of the system.

However, differently from [3], we concentrate here on a different characterization of the mode transition model concerning uncertain temporal information. We assume time to be *discrete*, and use natural numbers to denote time points. More importantly, we extend the basic assumption concerning the probability distribution of the lifetime of a component in the usual life phase to the different behavioral modes of the component. This means that we assume a memoryless distribution of the time spent by a component in a given mode, so the state transition model will be a DTMC. Let us discuss in more detail the representational issues; we assume that each component of the system has a set of possible behavioral modes, (one of which is the "correct" mode). The modes of a component are mutually exclusive with respect to a given time point. The fact that a component $c$ is in behavioral mode $m$ at time $t$ is represented by the atom $m(c,t)$. The relationships between behavioral modes and their manifestations are expressed as Horn clauses and the model is atemporal (see [3] for more details). Temporal information is abstracted from the logical model and is represented in a stochastic way. Let $COMPS$ be the set of components of the system to be diagnosed.

[2] It is possible to generalize to the continuous case (exponential lifetime) by considering the *failure rate* instead of the failure probability.

**Definition 3** *The* **Associated Markov Chain** *$AMC(c)$ of a component $c \in COMPS$ is a Markov chain whose states represent behavioral modes of c.*

It is clear that, assuming time to be discrete each $AMC$ will be a DTMC. Furthermore, if $p_i$ is the probability of being in the current mode $m_i$ at the next time instant, then the sojourn time $S_i$ in $m_i$ is geometrically distributed with parameter $(1-p_i)$ i.e. $P(S_i = t) = p_i^{t-1}(1-p_i)$. Notice that, assuming that a component has several fault modes and modeling the transitions among such modes as a Markov chain allows us to generalize the concept of failure probability to the concept of transition probability from one mode to another.

**Definition 4** *Let $AMC(c)$ be the Associated Markov Chain of a component $c \in COMPS$ and $p_{m_i m_j}(c)$ be the transition probability from mode $m_i$ to mode $m_j$ of the component c; the matrix* $\mathbf{P_c} = [p_{m_i m_j}(c)]$ *is called the* **transition probability matrix** *of the chain.*

The matrix $\mathbf{P_c}$ represents the transition probabilities from one mode to another in one time instant. The probabilistic behavior of a Markov chain is completely determined by its transition probability matrix and the initial probability distribution. In fact, let $\pi_c(n) = \{p^n_{m_1}(c), p^n_{m_2}(c), ...\}$ be the vector representing the probabilities of the states (modes) of the chain $AMC(c)$ at the instant $n$ ($p^n_{m_i}(c)$ is the probability of the component c being in mode $m_i$ at time $n$); this distribution depends on the initial distribution of modes, indeed the fundamental relation among mode probability distribution is given by

$$\pi_c(n) = \pi_c(0)\mathbf{P_c}^n$$

where the ($nth$ power) matrix $\mathbf{P_c}^n = [p^n_{m_i m_j}(c)]$ and $p^n_{m_i m_j}(c)$ is the probability of reaching mode $m_j$ from mode $m_i$ in $n$ time instants of the component $c$. Moreover, an interesting characteristic of Markov chains is the possibility of classifying its states in a rigorous way.

- an *ergodic set* of states is a set in which every state can be reached from every other state and which cannot be left once entered; each element of the set is called *ergodic state*;
- a *transient set* of states is a set in which every state can be reached from every other state and which can be left ; each element of the set is called *transient state*;
- an *absorbing state* is a state which once entered is never left (i.e. the probability of remaining in this state once entered is 1);

An interesting way of exploiting this classification is to use the $AMC(c)$ to give an *a-priori* classification of faults[3]. Given the $AMC(c)$, let $m_i$ be a fault mode of $c$:

- $m_i$ is a *permanent fault* iff it corresponds to an absorbing state of $AMC(c)$;
- $m_i$ is a *transient fault* iff it corresponds to a transient state of $AMC(c)$;
- $m_i$ is a *reversible fault* iff $p^n_{m_i m_0}(c) > 0$ for some $n$ and $m_0$ is the correct mode;
- $m_i$ is an *irreversible fault* iff $p^n_{m_i m_0}(c) = 0$ for every $n$ and $m_0$ is the correct mode.

## 3.2 CHARACTERIZATION OF A TEMPORAL DIAGNOSTIC PROBLEM

In order to show how stochastic information can be integrated in a logical model based diagnostic framework, in this section we briefly discuss a possible characterization of a temporal diagnostic problem. We can define a *temporal diagnostic problem* TPD as composed by a behavioral model $BM$, a set of components $COMPS$ and a set of observations at different time instants $OBS$ to be explained[4]; we can extract from a TPD a corresponding *atemporal diagnostic problem* APD by considering a set of observations $OBS(t)$ at a given time instant $t$. Solving an atemporal diagnostic problem at the instant $t$ means to determine an assignment at time $t$, $W(t)$, to each component in $COMPS$ explaining the observations in $OBS(t)$. This assignment is an *atemporal diagnosis*. Obviously, the time instants which are of interest in the diagnostic process are just those for which we have some observations; we will call such instants *relevant time instants*. Let us assume, as in the most part of the previous work on diagnosis, that each component is *independent* of each other (i.e.

[3] A similar proposal is presented in [11] by introducing an *a-posteriori* classification.

[4] The suitable notion of explanation can be extracted from the spectrum of definitions in [6]; moreover, notice that, for the sake of simplicity, we do not discuss the role of contextual information in a diagnostic problem (see [6] for an analysis of this problem).

the behavior of a component cannot influence the behavior of the others). We indicate as $m^c_{W(t)}$ the mode that $W(t)$ assigns to $c$; because of the independence of the components, if $t_i < t_j$ are two relevant time instant, we have

$$P[W(t_j)|W(t_i)] = \prod_{c \in COMPS} p_{m^c_{W(t_i)}, m^c_{W(t_j)}}(c)$$

and clearly

$$P[W(t)] = \prod_{c \in COMPS} p^t_{m^c_{W(t)}}(c)$$

where $p^t_{m^c_{W(t)}}(c)$ depends on the initial distributions $\pi_c(0)$ given for each component $c$. Moreover, we are also interested in the *joint probability* $P[W(0), W(1), \ldots W(t)]$ at time $t$ of a whole evolution $\{W(0) \ldots W(t)\}$. Since the analysis of the probabilities at time $t$ requires knowledge about the initial distribution of modes, in the absence of specific information we can reasonably assume an *uniform distribution*.

**Example 1:** let us suppose to have a simple system composed of a water pump P and a container C receiving the pumped water. For the sake of brevity we do not describe the logical behavioral model of this simple system, so we will suppose to have directly at disposal the atemporal diagnoses at the relevant instants. The AMCs of the two components and the transition probability matrices are shown respectively in figure 1 and in figure 2. Suppose to have the following hypothesis at time $t = 0$:

$W_1(0) = \{correct(P, 0), correct(C, 0)\}$
$W_2(0) = \{partially_occluded(P, 0), correct(C, 0)\}$
$W_3(0) = \{occluded(P, 0), correct(C, 0)\}$

By assuming an uniform distribution we get $P[W_1(0)] = P[W_2(0)] = P[W_3(0)] = \frac{1}{3}$. This means that we have the following initial distribution on the components (we assume the probabilities given in the following order: *punctured, leaking, correct* for C and *broken, occluded, leaking, partially_occluded, correct* for P).

$\pi_C(0) = (0, 0, 1)$
$\pi_P(0) = (0, \frac{1}{3}, 0, \frac{1}{3}, \frac{1}{3})$

Let us introduce the following abbreviations we will use in the examples: *punctured=pu, leaking=le, correct=co, broken=br, occluded=oc, partially_occluded=po*. Suppose that at time $t = 1$ the logical model concludes the atemporal diagnosis $W(1) = \{occluded(P), correct(C)\}$, then we have the following conditional probabilities

$P[W(1)|W_1(0)] = p_{co,oc}(P)\ p_{co,co}(C) = 0$

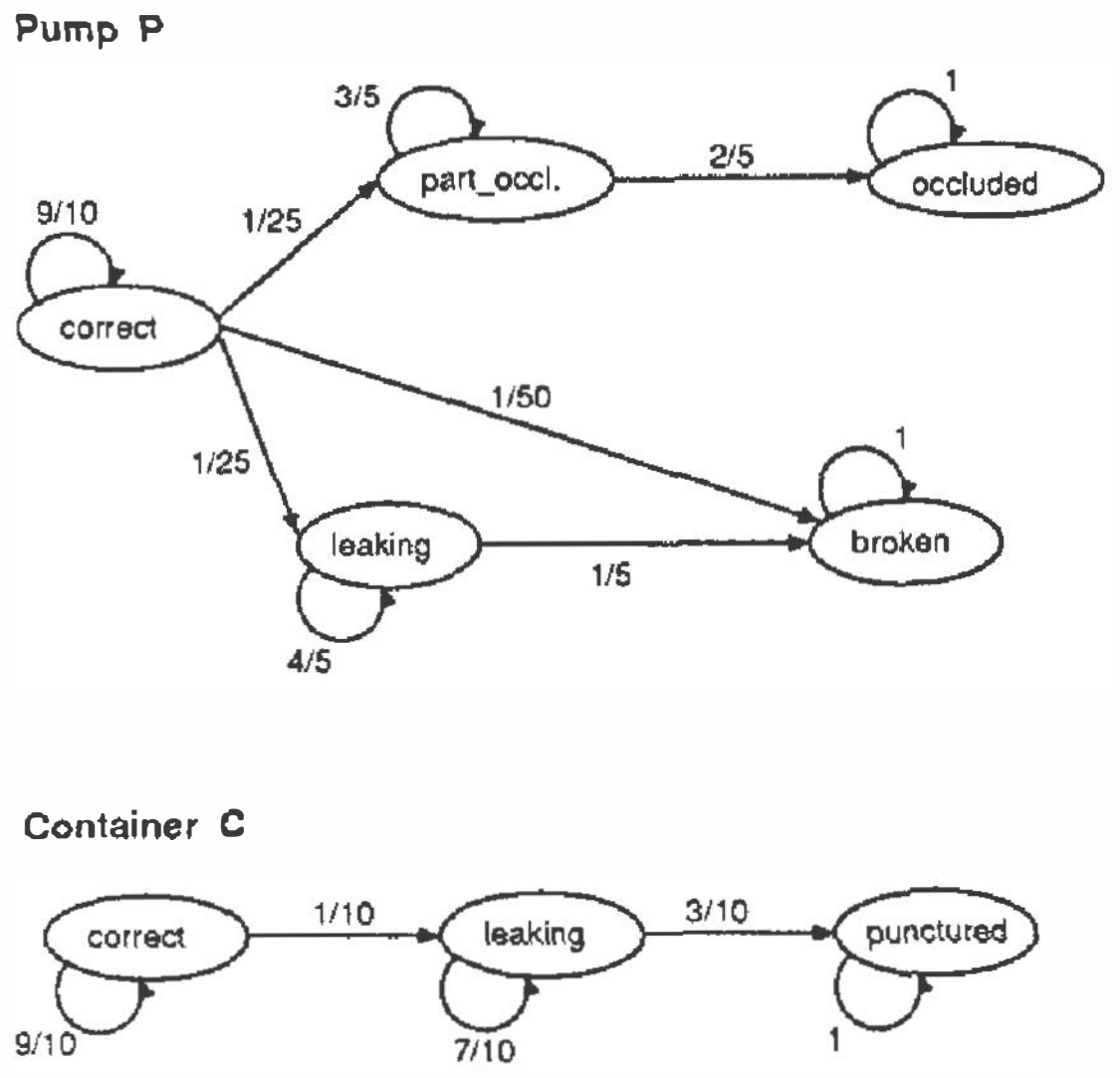

Figure 1: ACMs of the components P and C

$P[W(1)|W_2(0)] = p_{po,oc}(P)\ p_{co,co}(C) = \frac{9}{25}$
$P[W(1)|W_3(0)] = p_{oc,oc}(P)\ p_{co,co}(C) = \frac{9}{10}$

and the following joint probabilities

$P[W_1(0), W(1)] =$
$\quad P[W_1(0)]P[W(1)|W_1(0)] = \frac{1}{3}0 = 0$
$P[W_2(0), W(1)] =$
$\quad P[W_2(0)]P[W(1)|W_2(0)] = \frac{1}{3}\frac{9}{25} = \frac{3}{25}$
$P[W_3(0), W(1)] =$
$\quad P[W_3(0)]P[W(1)|W_3(0)] = \frac{1}{3}\frac{9}{10} = \frac{3}{10}$

It is now possible to define a notion of temporal diagnosis on the basis of the concept of solution for an atemporal problem.

**Definition 5** *Given an atemporal diagnostic problem APD, an assignment $W(t)$ of behavioral modes to each $c \in COMPS$ is a* **temporal diagnosis** *for the corresponding temporal diagnostic problem iff:*

1. *$W(t)$ is a solution for ADP for every relevant time instant $t$;*

2. *for every consecutive relevant time instants $t_i < t_j$, $P[W(t_j)|W(t_i)] \geq \sigma$ where $0 \leq \sigma \leq 1$ is a predefinite threshold named the* **plausibility threshold** *for the temporal diagnostic problem.*

Notice that the second part of the definition could be different if we supposed to have information on the plausibility threshold directly on the com-

| | broken | occluded | leaking | part_occl. | correct |
|---|---|---|---|---|---|
| broken | 1 | 0 | 0 | 0 | 0 |
| occluded | 0 | 1 | 0 | 0 | 0 |
| leaking | 1/5 | 0 | 4/5 | 0 | 0 |
| part_occl. | 0 | 2/5 | 0 | 3/5 | 0 |
| correct | 1/50 | 0 | 1/25 | 1/25 | 9/10 |

$P_P$ : transition probability matrix for component **P**

| | punctured | leaking | correct |
|---|---|---|---|
| punctured | 1 | 0 | 0 |
| leaking | 3/10 | 7/10 | 0 |
| correct | 0 | 1/10 | 9/10 |

$P_C$ :transition probability matrix for component **C**

Figure 2: Transition probability matrices of the components P and C

ponents; in this case we could check the plausibility of the transitions of each component $c$ from $r = m^c_{W(t_i)}$ to $s = m^c_{W(t_j)}$. Obviously, in this case the corresponding threshold should be different; indeed, given the same threshold $\sigma$, if $P[W(t_j)|W(t_i)] \geq \sigma$ and $n = t_j - t_i$, then $p^n_{rs} \geq \sigma$ but not vice versa.
The information in the Markov chains associated with the components can then be used to discard very improbable diagnostic hypotheses from the temporal point of view.

**Example 2:** given the same system of example 1, suppose we get the atemporal diagnosis $W(0) = \{correct(P,0), correct(C,0)\}$. Consider now the case in which new observations allow us to conclude, at time $t = 1$ the following diagnostic hypotheses:

$W_1(1) = \{occluded(P,1), correct(C,1)\}$
$W_2(1) = \{broken(P,1), correct(C,1)\}$
$W_3(1) = \{correct(P,1), punctured(C,1)\}$

Let the plausibility threshold be $\sigma = \frac{1}{100}$. We have

$P[W_1(1)|W(0)] = P[W_3(1)|W(0)] = 0$
$P[W_2(1)|W(0)] = \frac{1}{50}\frac{9}{10} = \frac{9}{500} > \sigma$

The only admissible diagnosis at time $t = 1$ is $W_2(1)$. By computing the square matrices ${\mathbf{P_P}}^2$ and ${\mathbf{P_C}}^2$, it is easy to see that, if the same diagnoses were concluded at time $t = 2$, then the only non-admissible diagnosis would be the first one.

It should be clear that an interesting aspect of this approach is the possibility of ranking the temporal diagnoses we get; indeed, it is easy to show that the joint probabilities at time $t$ can be recursively computed, given the a-priori distributions $P[W(0)]$ by the following formula:

$$P[W(0), ...W(t-k), W(t)] = P[W(0), ...W(t-k)]P[W(t)|W(t-k)]$$

where the first factor is the joint probability at the previous step and the second can be computed by means of the matrices ${\mathbf{P_C}}^k$. For instance, in example 1, the most probable temporal diagnosis is $\{W_3(0), W(1)\}$.

# 4 DISCUSSION

In this paper, an approach integrating stochastic information in logical model-based diagnosis is proposed, in the attempt to characterize problems deriving from the introduction of temporal information in the model of the system to be diagnosed. However, an important problem still to be considered concerns the role of the *AMC*'s of the components with respect to the underlying static logical model. In fact, using the approach described in this paper, we assume to model the temporal behavior of the components at a very abstract point of view in which a transition from one mode to another is just a random function of time. The problem arises when stochastic information contrasts with information gathered by means of new observations on the system. Actually, it is reasonable to assume that new observations supply new evidence for the atemporal diagnostic hypotheses at a given instant $t$, even if the predicted probabilities were low. The problem is to figure out how to combine this new evidence obtained by the logical step with the predictions gathered from the stochastic model; this means that we have to perform a revision of the probabilities at time $t$ in order to take into account the new information. However, we have to paid attention to the context in which such a revision is reasonable. The assumption we make is that the reasoning step on the logical model cannot loose any possible diagnosis; so if $W_1(t) \ldots W_n(t)$ are the atemporal diagnoses at time $t$, there not exists any other possible diagnosis at that time, even if some of them could not actually be diagnoses (the observations might not allow us to discriminate between them). In this context, the probabilities of the temporal diagnoses at time $t$ can be normalized to sum 1. If the number of joint probabilities at time $t$ is $N$ and we indicate as $P_i(t)$ the $ith$ of such probabilities, the normalization factor at time $t$ will be

$$F(t) = \frac{1}{\sum_{i=1}^{N} P_i(t)}$$

We can now compute the revised joint probabilities as $RP_i(t) = P_i(t)F(t)$, the revised conditional ones as $RP[W(t)|W(t-k)] = P[W(t)|W(t-k)]F(t)$ and we can check the plausibility threshold on these new values. Notice that if we check the plausibility threshold on the transition probabilities of the components, we have to compute, at each relevant instant $t$, for each component $c$, the distribution $\pi_c(t) = \pi_c(t-k)\mathbf{P_c}^k$ which is not needed in the previous case. Indeed, in this case, we need to normalize the probabilities of each component in a given mode using a factor $f(c,t)$. In this way we can get the revised transition probabilities from mode $m_i$ to $m_j$ of $c$ as $rp^k_{m_i m_j} = p^k_{m_i m_j} f(c,t)$ and checking the plausibility threshold on this value.

**Example 3:** Consider the situation at time $t = 0$ and $t = 1$ in example 1 when $W(1)$ is the only atemporal diagnosis at time $t = 1$.

$F(1) = \frac{1}{\frac{3}{25}+\frac{3}{10}} = \frac{50}{21}$

$RP[W(1)|W_1(0)] = 0$ $RP[W(1)|W_2(0)] = \frac{6}{7}$ $RP[W(1)|W_3(0)] = \frac{15}{7}$

If we take trace of the mode probabilities of the components we will have

$\pi_C(1) = (0, \frac{1}{10}, \frac{9}{10})$
$\pi_P(1) = (\frac{1}{50}, \frac{7}{15}, \frac{1}{75}, \frac{16}{75}, \frac{3}{10})$

Since at $t = 1$ $C$ can only be *correct* and $P$ *occluded*, we get $f(C,1) = \frac{10}{9}$ and $f(P,1) = \frac{15}{7}$. So, for example the revised transition probabilities from *partially_occluded* to *occluded* of $P$ will be

$rp_{po,oc}(P) = p_{po,oc}(P)f(P,1) = \frac{6}{7}$

and that from *correct* to *correct* of component $C$ will be

$rp_{co,co}(C) = p_{co,co}(C)f(C,1) = 1.$

Notice that performing the revision means that the conclusions obtained from the static model are more important than the probabilistic predictions of the stochastic models of the components; this can happen, for instance, when the relations between behavioral modes and their manifestations can be safely abstracted from time and we have all the observations, or at least the most important ones, at disposal. In such a case, the assumption of getting from the logical model all the possible atemporal diagnostic hypothesis can be considered suitable and reasonable.

We did not discussed here problems concerning reasoning issues, however considerations similar to those discussed in [3] can be applied to the framework described in the present paper.

### Acknowledgements

The author would like to thank P. Torasso and L. Console for their valuable help and useful discussions about the topic presented in the paper. This work has been partially supported by CNR Temporal Reasoning Project under grant n. 91.02351.CT12.